\pdfoutput=1

\documentclass[11pt]{article}

\usepackage{acl}

\usepackage{times}
\usepackage{latexsym}
\usepackage{multirow}
\usepackage{graphicx} 
\usepackage{enumitem}
\usepackage{placeins}
\usepackage{comment}
\usepackage{caption}
\usepackage{subcaption}

\usepackage[T1]{fontenc}

\usepackage[utf8]{inputenc}

\usepackage{microtype}

\usepackage{inconsolata}

\usepackage{booktabs}

\newcommand{\mee}[1]{\textcolor{blue}{AA: #1}}

\title{Exploring the Maze of Multilingual Modeling}

\author{Sina Bagheri Nezhad, Ameeta Agrawal \\
  Portland State University \\
  \texttt{\{sina.bagherinezhad, ameeta\}@pdx.edu} \\}

\begin{document}
\maketitle
\begin{abstract}
Multilingual language models have gained significant attention in recent years, enabling the development of applications that meet diverse linguistic contexts. In this paper, we present a comprehensive evaluation of three popular multilingual language models: \texttt{mBERT}, \texttt{XLM-R}, and \texttt{GPT-3}. We assess their performance across a diverse set of languages, with a focus on understanding the impact of resource availability (general and model-specific), language family, script type, and word order on model performance, under two distinct tasks -- text classification and text generation. Our findings reveal that while the amount of language-specific pretraining data plays a crucial role in model performance, we also identify other factors such as general resource availability, language family, and script type, as important features. We hope that our study contributes to a deeper understanding of multilingual language models to enhance their performance across languages and linguistic contexts.
\end{abstract}

\section{Introduction}

Multilingual language models have transformed natural language processing (NLP) by enabling applications such as machine translation and sentiment analysis in multiple languages. 
Continuous efforts are dedicated to understanding of multilingual models' performance across languages with distinct linguistic properties \cite{devlin-etal-2019-bert, wu2020all, scao2022bloom, lai2023chatgpt, ahuja2023mega}. Despite several efforts, linguistic disparity in NLP persists \cite{joshi-etal-2020-state,ranathunga2022some}. It remains important to to not only improve the performance of the models for most languages of the world, but also to make them safer by focusing on alignment beyond English \cite{wang2023all}.


However, it remains unclear which factors truly contribute to the development of effective multilingual models. Several studies indicate the amount of language-specific data available in the pretraining corpus as one of the key factors \cite{wu2020all}. However, most studies are conducted for a limited set of languages on a given task, focusing on a limited set of training paradigm (such as masked language modeling (MLM) or autoregressive), and especially on a handful of factors.

In this work, we contribute to this area of research by comprehensively evaluating three multilingual language models of type MLM and autoregressive (\texttt{mBERT} \cite{devlin-etal-2019-bert}, \texttt{XLM-R} \cite{conneau-etal-2020-unsupervised} and \texttt{GPT-3} \cite{brown2020language}) under two types of tasks (text classification and text generation) covering a wide range of languages. More imporantly, we consider five different factors in our analysis (pretraining data size, general resource availability levels, language family, script type, and word order). We leverage the recently introduced SIB-200 dataset as well as create a novel multilingual dataset of recently published BBC news articles in 43 languages, called \texttt{mBBC}, which allows us to evaluate on text that may not have been seen by these models during their training.



Through an extensive multivariate and univariate analysis, we find that while model-specific resource availability strongly influences model performance in certain cases, this does not appear to be true for all models and all tasks. Other factors identified as important include general resource availability, language family, and script type.


We hope that our findings will help researchers and practitioners to develop more inclusive and effective multilingual NLP systems.


    


\begin{table*}[ht]
\centering
\resizebox{\textwidth}{!}{
\begin{tabular}{l|p{8cm}|p{3.5cm}|c}
\hline
\textbf{Reference} & \textbf{Factors} & \textbf{Task} & \textbf{Languages} \\
\hline
\citet{wu2020all} & Pretraining data size, Task-specific data size, Vocabulary size &  NER & 99 \\
\hline
\citet{scao2022bloom} & Pretraining data size, Task-specific data size, Language family, Language script  & Probing & 17\\
\hline
\citet{shliazhko2022mgpt} & Pretraining data size, Language script, Model size & Perplexity & 61 \\
\hline
\citet{ahuja2023mega} & Pretraining data size, Tokenizer fertility  & Classification, QA, Sequence Labeling, NLG, RAI & 2-48 \\
\hline
Ours & Pretraining data size, Language family, {\bf Language script,  General resource availability, Word order} & Text classification, Text generation & 204, 43 \\
\hline
\end{tabular}
}
\caption{Factors considered in related works and this work. Factors distinct to our work are shown in bold.}
\label{table:related_works}
\end{table*}

\section{Related Work}

Multilingual NLP research has made significant strides, introducing the development and evaluation of several multilingual language models trained on diverse and combined language datasets 
 (\texttt{mBERT} \cite{devlin-etal-2019-bert}, \texttt{XLM-R} \cite{conneau-etal-2020-unsupervised}, \texttt{mBART} \cite{10.1162/tacl_a_00343}, \texttt{mT5} \cite{xue-etal-2021-mt5}, \texttt{BLOOM} \cite{scao2022bloom}, \texttt{GPT-3} \cite{brown2020language}, \texttt{GPT-4} \cite{openai2023gpt4}, \texttt{LLaMA} \cite{touvron2023llama}, \texttt{PaLM} \cite{chowdhery2022palm}, \texttt{PaLM 2} \cite{anil2023palm}, and others). 


Factors that may have an impact on the performance of multilingual models are being increasingly investigated. \citet{wu2020all} used the named entity recognition task and considered three factors that might affect the downstream task performance: pretraining data size, task-specific data size, and
vocabulary size in task-specific data. They found that the larger the task-specific supervised dataset, the better the downstream performance on NER. \citet{scao2022bloom} studied the correlation between probing performance and several factors, and found that the results of BLOOM-1B7 are highly correlated
with language family, task-specific dataset size, and pretraining dataset size. \citet{shliazhko2022mgpt} used perplexity to assess the impact of language script, pretraining corpus size, and model size, and found that the language modeling performance depends on the model size and the pretraining corpus size in a language, whereas \citet{ahuja2023mega} studied the impact of tokenizer fertility and pretraining data, and found that the models perform worse in languages for which the tokenizer is of poor quality, and that the amount of training data available in a language can partially explain some results. 


In contrast, we conduct a more holistic investigation to provide better insights related to three multilingual language models (both MLM and autoregressive) across two distinct tasks (a supervised task such as text classification, and an unsupervised text generation task). Moreover, prior work  studied only a few languages for a given task primarily because of limited availability of annotated datasets. 
The recent landscape of multilingual datasets, however, has seen remarkable contributions \cite{costa2022no, adelani2023sib,imanigooghari-etal-2023-glot500}, offering valuable resources for diverse linguistic analysis. 
While these resources are used in our analysis, we further create mBBC to support  unsupervised modeling, encompassing news from 2023 in 43 languages. Concerns of data contamination remain persistent \cite{golchin2023time, deng2023investigating} and using mBBC ensures that the evaluation uses data that was unseen by the language models considered in our study. Moreover, it addresses the need for a dataset that can be leveraged without fine-tuning language models, mitigating the impact of hyperparameter tuning in our analytical pursuits. Table~\ref{table:related_works} presents an overview of some of the related works.

\section{Exploring the Maze of Multilingual Modeling}

Several factors can influence the performance of multilingual models. 
In this study, we consider three multilingual models, five distinct factors related to typology and data, and two types of NLP tasks.

\subsection{Models}
The three multilingual language models studied in our analysis include \texttt{mBERT} (bert-base-multilingual-cased) \cite{devlin-etal-2019-bert}, \texttt{XLM-R} (xlm-roberta-base) \cite{conneau-etal-2020-unsupervised}, and \texttt{GPT-3} (text-davinci-003) \cite{brown2020language}. \texttt{mBERT} and \texttt{XLM-R} are masked language models, while \texttt{GPT-3} is an autoregressive language model. These models were selected because of their extensive language support, allowing us to maximize the linguistic diversity covered in our analysis.  Additionally, the choice of \texttt{mBERT} and \texttt{XLM-R} was influenced by the fact that these models, after fine-tuning, continue to demonstrate competitive performance, even rivaling larger language models such as \texttt{ChatGPT} \cite{lai2023chatgpt,zhu2023multilingual}. 


\subsection{Typology and Data Factors}
We consider various factors to understand their impact on model performance including:

\begin{itemize}[leftmargin=*]

 \item \textbf{Pretraining Data Size (Train Token (TT)):} This is the amount of language-specific pretraining data (million tokens) used by each model during training\footnote{We obtained the Train Token (TT) values for \texttt{mBERT} from \url{https://github.com/mayhewsw/multilingual-data-stats}, for \texttt{XLM-R} from its paper \cite{conneau-etal-2020-unsupervised}, and for \texttt{GPT-3} we use proxy statistics from \url{https://github.com/openai/gpt-3/blob/master/dataset_statistics/languages_by_word_count.csv}.}. 

  \item \textbf{General Resource Availability (Res Level):} Beyond model-specific resources such as pretraining data size, we also consider a more general notion of resource availability, as per the linguistic diversity taxonomy which categorizes languages into six resource levels \cite{joshi-etal-2020-state}. This classification helps us understand the influence of more general resource availability on model performance, and may serve as a proxy when model-specific statistics may not be available (such as in the case of commercial models).

  \item \textbf{Language Family (Lang Family):} The language families that the languages belong to capture some of their linguistic relationships. The information was sourced from the Ethnologue\footnote{\url{https://www.ethnologue.com}} \cite{Ethnologue_2022}.

  \item \textbf{Script:} The script of a language refers to the writing system it employs. This information was sourced from ScriptSource\footnote{\url{https://www.scriptsource.org}}.

  \item \textbf{Word Order:} Word order refers to the arrangement of syntactic constituents within a language. This feature captures the structural variations in how languages express relationships between subject, object, and verb (e.g., Subject-Object-Verb (SOV), Subject-Verb-Object (SVO), and Verb-Subject-Object (VSO)). This information was sourced from \citet{wals}.

\end{itemize}


\subsection{Tasks and Datasets}
We systematically study the multilingual models under two distinct and important tasks -- text classification and text generation \cite{chang2023language}.


\medskip

\noindent \textbf{Text Classification on SIB-200 dataset} \quad The SIB-200 dataset \cite{adelani2023sib} facilitates the text classification task in 204 languages, where each instance of text is categorized into one of six classes. The performance is measured in terms of F1 score.


The \texttt{mBERT}  and \texttt{XLM-R}  models were fine-tuned on the training set of SIB-200 and evaluated on a separate test set. The \texttt{GPT-3}  model was used under the zero-shot setting without any specific fine-tuning. Default train and test splits with hyperparameters introduced by the authors of SIB-200 were used. 



\medskip
\noindent \textbf{Text Generation on mBBC dataset} \quad As autoregressive models have become increasingly popular, so has the task of text generation, where the models select each next token given some context. Such a task presents a complemetary way of evaluation by not requiring any labeled data. Given a sequence of $n$ tokens, the models predict the next token $n+1$. We formulate this as a binary classification task: if the ground truth token matches any token in the top $k$ predicted tokens generated by the models, then the output is considered to be correct\footnote{We experimented with various hyperparameter settings and finally empirically set $n=30$ and $k=5$.}. The results are reported in terms of accuracy. For each model, we utilize their respective tokenizers to preprocess the input sequences. For each language in mBBC, we experiment with 2000 samples, which allows us to obtain statistically significant results while ensuring computational feasibility. The experimental procedure and implementation details are described in Appendix~\ref{app:procedure}.




\begin{figure*}[ht]
\centering
\includegraphics[width=1\textwidth]{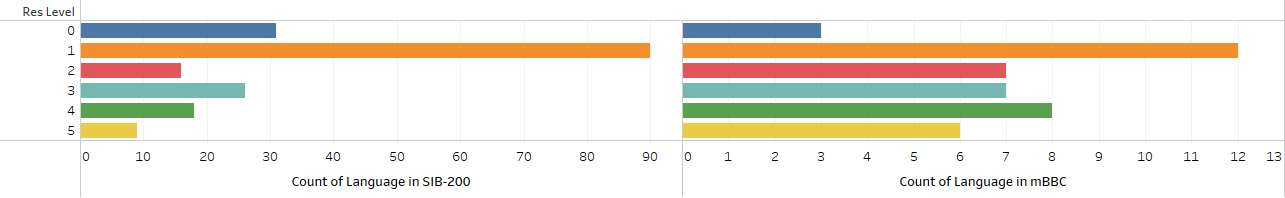}
\caption{Distribution of resource level in SIB-200 and \texttt{mBBC} datasets.}
\label{figure:reslevel}
\end{figure*}

\begin{figure*}[ht]
\centering
\includegraphics[width=1\textwidth]{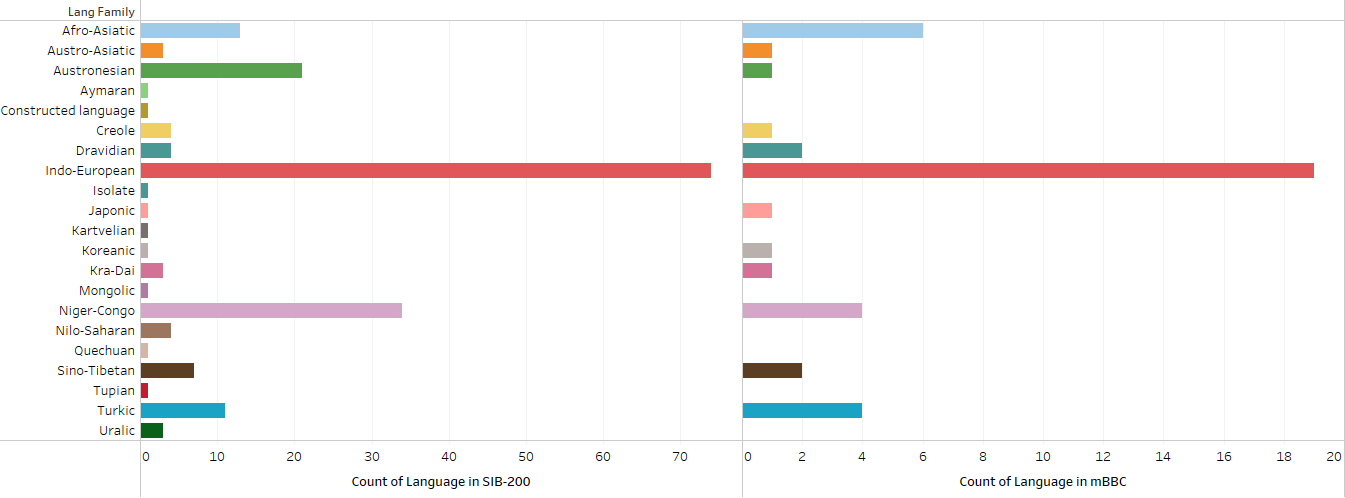}
\caption{Distribution of language family in SIB-200 and \texttt{mBBC} datasets.}
\label{figure:lang}
\end{figure*}

\begin{figure*}[!t]
     \centering
     \begin{subfigure}[b]{0.9\textwidth}
         \centering
         \includegraphics[width=\textwidth]{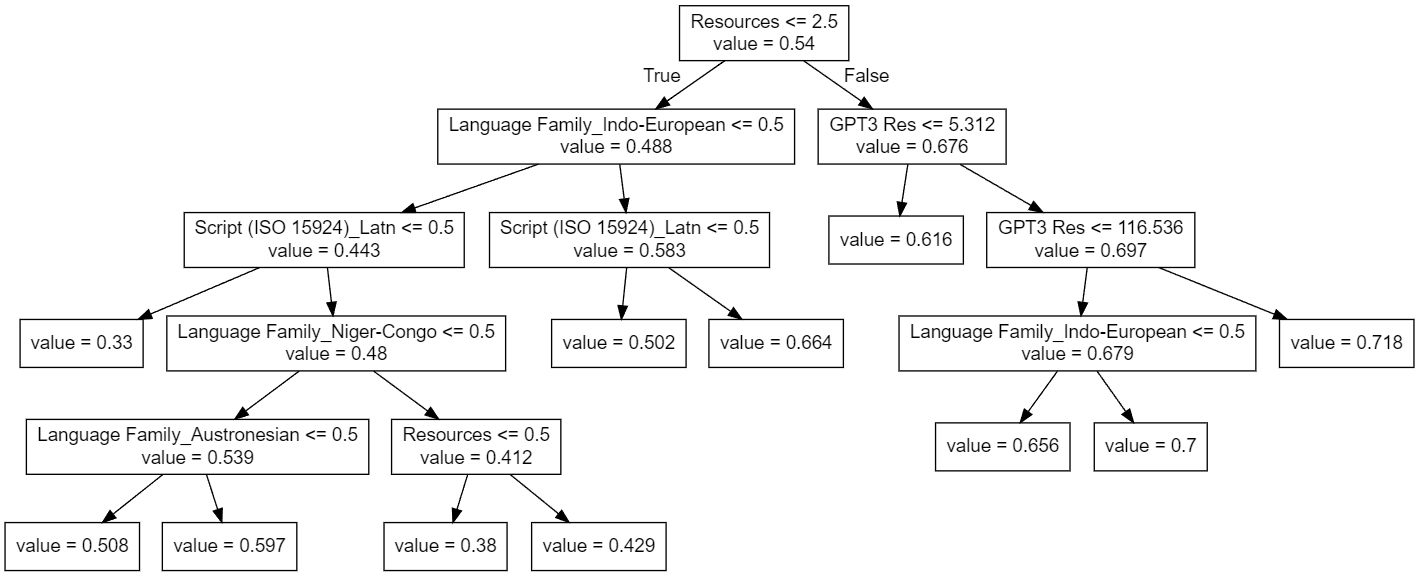}
         \caption{Decision tree visualization for \texttt{GPT-3} model on SIB-200 dataset}
         \label{a}
     \end{subfigure}\\
     
     \begin{subfigure}[b]{0.4\textwidth}
         \centering
         \includegraphics[width=\textwidth]{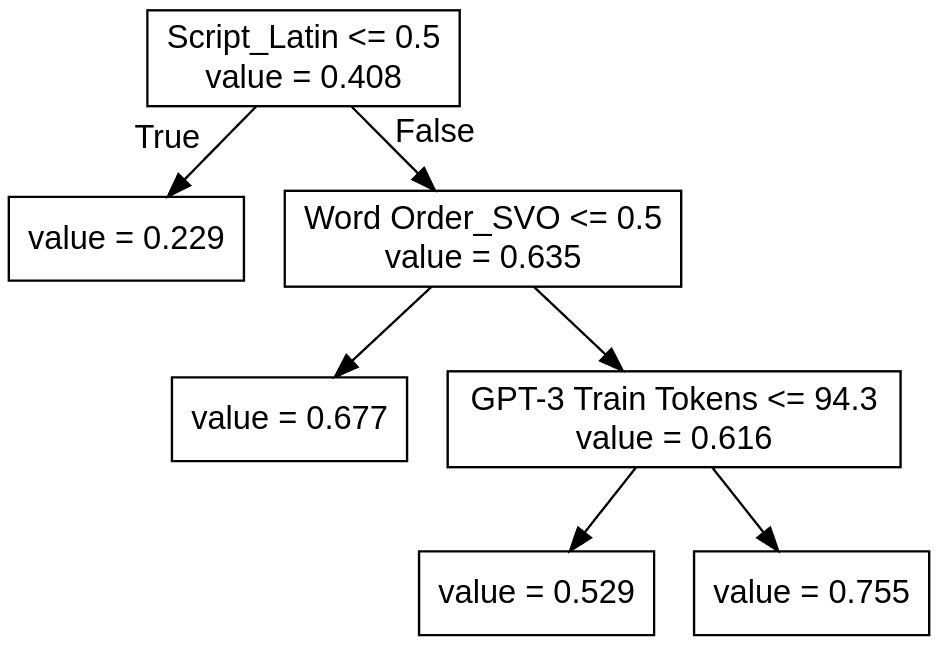}
         \caption{Decision tree visualization for \texttt{GPT-3} model on mBBC dataset}
         \label{b}
     \end{subfigure}
        \caption{Decision tree visualization. \emph{Value} refers to the expected F1 score/accuracy of the model. 
        }
        \label{fig:dt}
\end{figure*}

\medskip
\noindent \textbf{\texttt{mBBC} (multilingual BBC)} \quad To create this new multilingual news dataset,  news articles were gathered from BBC news in 43 different languages\footnote{\url{https://www.bbc.co.uk/ws/languages}}, which, in constrast to SIB-200, presents a relatively real-world snapshot of language distribution based on the fact that BBC broadcasts news in these 43 languages providing a global coverage. Most importantly, the articles are sourced from mid 2023 which allows us to be reasonably confident that the models considered in our study have not been exposed to this data during their training, thereby limiting concerns of data contamination. Additionally, by exclusively sourcing articles from a single source, consistency in tone and writing style across diverse languages is maintained, facilitating a more comparable evaluation.

The dataset includes languages from 12 language families and 16 scripts. Detailed statistics of  \texttt{mBBC} dataset's languages, including language family, script, and other relevant linguistic characteristics are presented in Appendix \ref{app:mBBC}. Among the languages available in \texttt{mBBC}, \texttt{mBERT} was able to support 32, while \texttt{XLM-R} supported 38, with 31 of them overlapping with \texttt{mBERT}'s supported languages. \texttt{GPT-3} was run on all 43 languages in our dataset.

\subsection{Analysis of SIB-200 and mBBC}

Figure~\ref{figure:reslevel} shows that most languages present in SIB-200 are classified as resource level 1, which is intentional by design. However, mBBC which was created by what was naturally available on the BBC website also contains a significant number of low resource languages, with the majority falling under resource level 1. This indicates that while linguistic resources may be limited for many languages, they are still utilized by communities and services such as BBC News in the real world, emphasizing the need for considerable attention to these underserved languages. 

Figure~\ref{figure:lang} shows that Indo-European languages dominate both datasets (about 36\% of SIB-200 and 44\% of mBBC), reflecting their status as the most widely spoken language family in the world \cite{Ethnologue_2022}. In SIB-200, the two other language families with considerable presence include Niger-Congo and  Austronesian, whereas in mBBC, it is Afro-Asiatic and Niger-Congo. 

In terms of writing systems, the Latin script is the most common across both the datasets, being used by nearly 70\% of the global population \cite{Vaughan_2020}. The next two most frequent scripts are Arabic and Cyrillic across both the datasets (see Figure \ref{figure:script} in Appendix \ref{app:mBBC}). 


\section{Results and Analysis}
In this section, we present the results of our evaluation of multilingual language models and analyze their performance based on various factors including resources (model-specific and general), language family, script, word order, and their interactions. 

\subsection{Multivariate Analysis}
To collectively analyze and understand the intricate interplay of multiple factors, which are of different types such as categorical, ordinal, and numeric, we use decision tree analysis for statistical inference to identify influential features. This was followed by Mann-Whitney U test \cite{mann1947test} for the classification task and Fisher's exact test \cite{fisher1922interpretation} for the generation task to determine significant differences. Decision trees are trained to predict the accuracy and F1 score of models based on language features, and thus, analyzing them allows us to gain insights into the significance of features.

Figure~\ref{fig:dt} presents the decision tree analysis of the \texttt{GPT-3} model for SIB-200 and mBBC datasets. Other results are included in Appendix \ref{app:charts} and \ref{app:SIB}. According to the analysis, for SIB-200, {\em general resource level} (more or less than 2.5) is identified as the most important feature. For lower resource languages (the left child node), language family is the next most important feature, whereas for higher resource languages (the right child node), the train token size is the next most important feature. For mBBC, {\em script type} (Latin or not) appears to be the most important feature. All results are statistically significant (p < 0.001).

\begin{figure*}
     \centering
     \begin{subfigure}[b]{\textwidth}
         \centering
         \includegraphics[width=\textwidth]{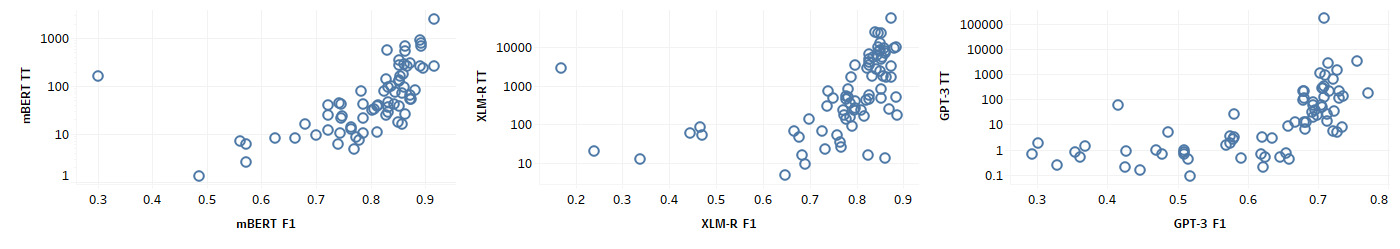}
         \caption{F1 Score vs. Train Token for SIB-200.}
         \label{a}
     \end{subfigure}\\
     
     \begin{subfigure}[b]{\textwidth}
         \centering
         \includegraphics[width=\textwidth]{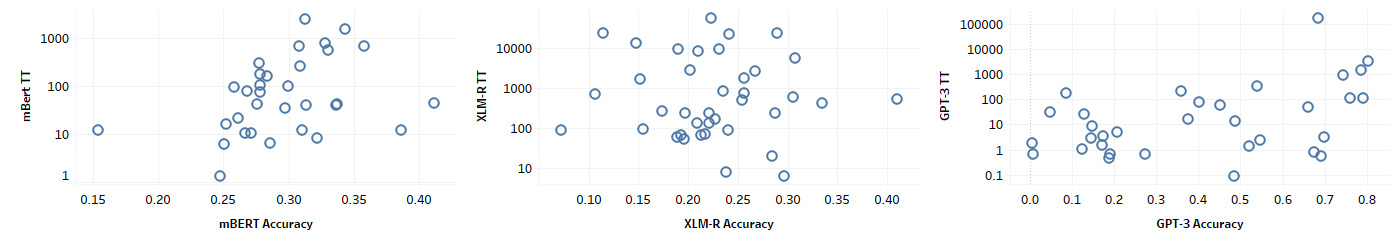}
         \caption{Accuracy vs. Train Token for \texttt{mBBC}}
         \label{b}
     \end{subfigure}
        \caption{Correlation analysis between performance and pretraining data (train tokens)}
        \label{fig:perf-TT}
\end{figure*}

\begin{figure*}[t]
     \centering
     \begin{subfigure}[b]{0.8\textwidth}
         \centering
         \includegraphics[width=\textwidth]{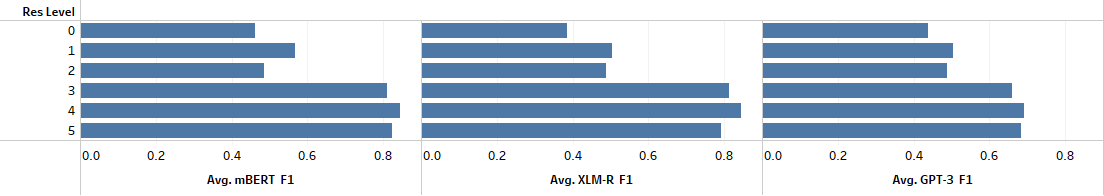}
         \caption{Average F1 score of  \texttt{mBERT}, \texttt{XLM-R}, and \texttt{GPT-3} across different resource levels on SIB-200.}
         \label{a}
     \end{subfigure}\\
     
     \begin{subfigure}[b]{0.8\textwidth}
         \centering
         \includegraphics[width=\textwidth]{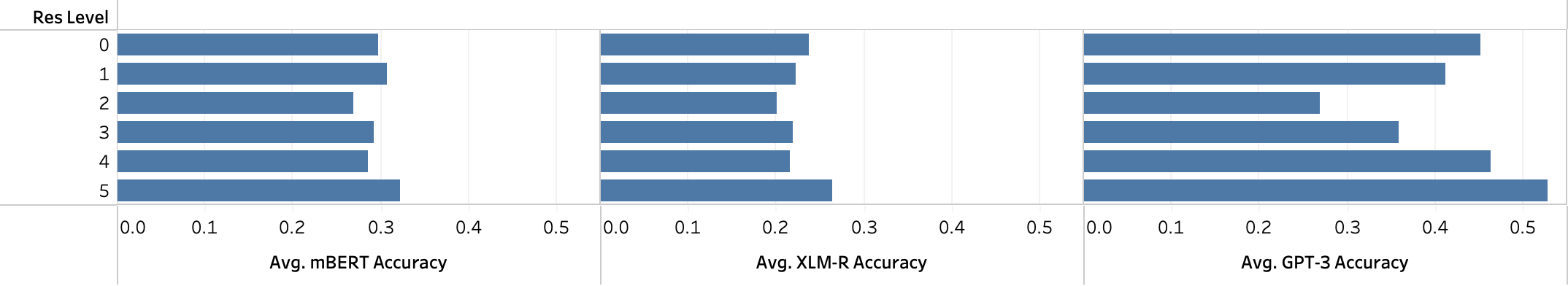}
         \caption{Average accuracy of \texttt{mBERT}, \texttt{XLM-R}, and \texttt{GPT-3} across different resource levels on \texttt{mBBC}}
         \label{b}
     \end{subfigure}
        \caption{Model results across different resource levels}
        \label{fig:perf-res}
\end{figure*}

\begin{figure*}[ht]
\centering
\includegraphics[width=0.9\textwidth]{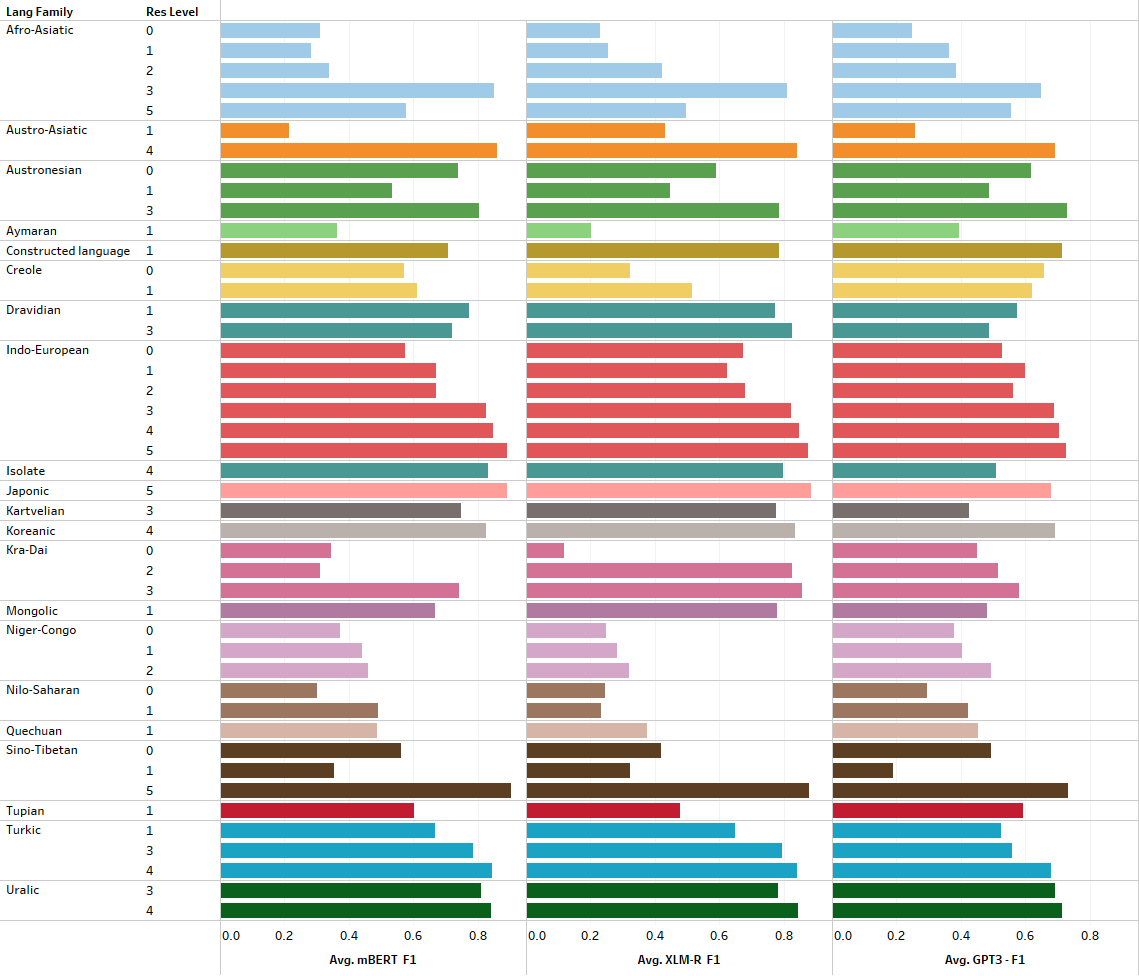}
\caption{Average accuracy of mBERT, XLM-R, and GPT-3  across language families and resource levels for text classification on SIB-200. The results
within each language family are averaged for all languages of the same resource levels}
\label{figure:SIB-lang-res}
\end{figure*}

\begin{figure*}[ht]
\centering
\includegraphics[width=0.8\textwidth]{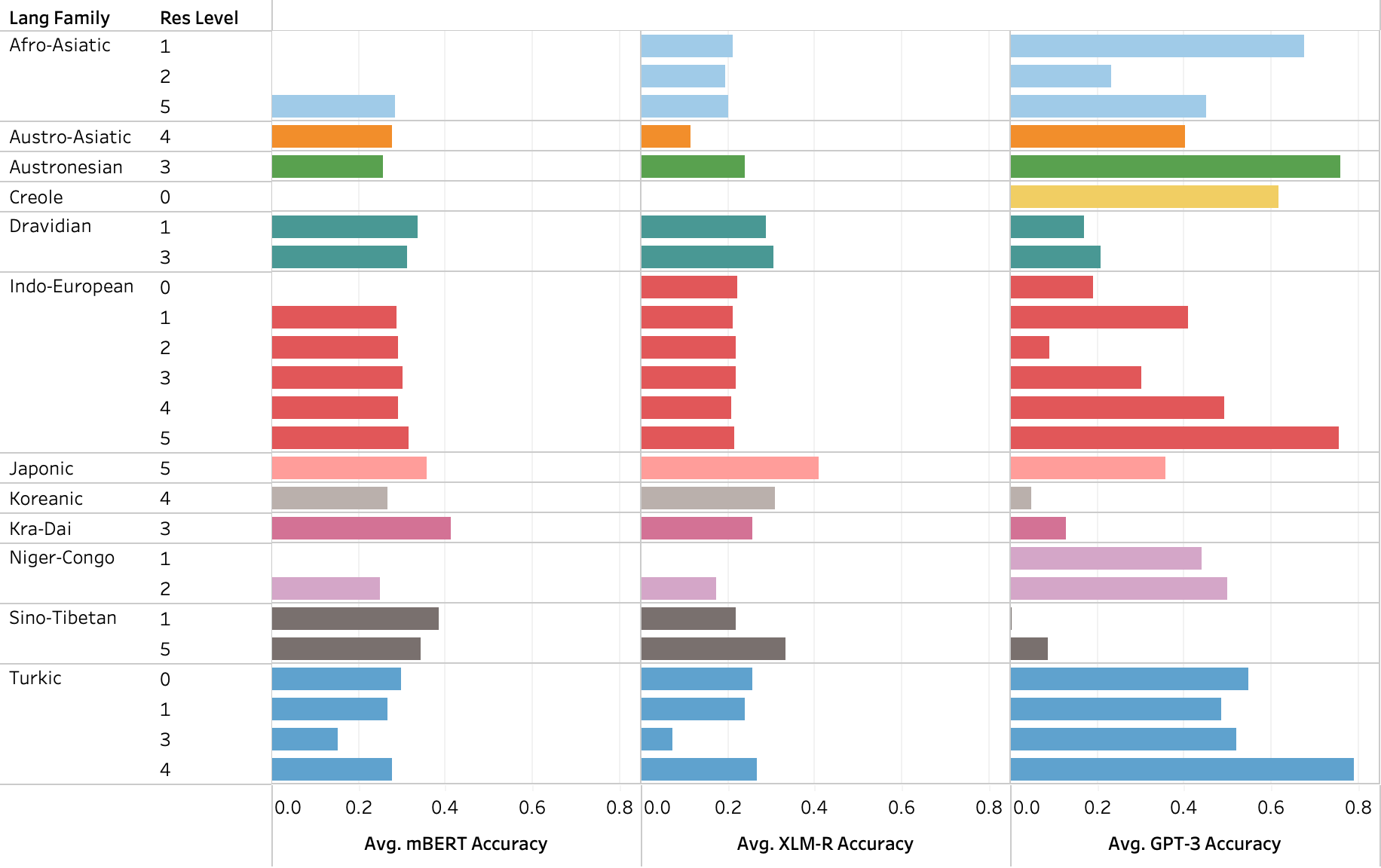}
\caption{Average accuracy of mBERT, XLM-R, and GPT-3 across language families and resource levels for text generation on mBBC. The results within each language family are averaged for all languages of the same resource levels. 
}
\label{figure:lang+res}
\end{figure*}

\begin{table}[h]
\centering
\begin{tabular}{l|c|c}
\cline{2-3}
\multicolumn{1}{c|}{} & \multicolumn{1}{c|}{\texttt{SIB-200}} & \multicolumn{1}{c|}{\texttt{mBBC}}\\ \hline
\texttt{mBERT} & Pretraining data & Language family \\ \hline
\texttt{XLM-R} & Pretraining data & Script type     \\ \hline
\texttt{GPT-3} & Resource Level   & Script type      \\ \hline
\end{tabular}
\caption{Top feature in decision trees. It shows for the downstream task, training size and resource level is the top feature while for text generation task linguistic characters are more important. The p-values for all features in this table are less than 0.001. The p-values for SIB-200 is calculated based on Mann-Whitney U test and p-values for mBBC is calculated by Fisher's exact test.}
\label{tab:trees}
\end{table}

Table~\ref{tab:trees} summarizes the results of all the decision tree analyses (full results are included in Appendix \ref{app:charts} and \ref{app:SIB}). In general, for text classification on SIB-200, two out of three models are most impacted by the  model-specific pretraining data size. However, general resource availability based on linguistic diversity taxonomy \cite{joshi-etal-2020-state} appears to be the most important factor for GPT-3. 

Interestingly, however, for text generation using mBBC dataset, the decision tree analysis reveals factors other than resource availability to be most important. For GPT-3 and XLM-R, it is script type, an often overlooked factor, whereas for mBERT, it is language family.

Taken together, these results suggest that there appear to be model-based as well as task-based differences that affect what is considered to be the most important factor in predicting a model's performance on a given task, and that in only 2 out of 6 settings (3 models x 2 tasks), pretraining data size was indicated as the most important factor, with general resource levels, script type, and language family also emerging as important factors in other settings. In other words, the same model may be impacted by different factors depending on the task at hand (classification vs. generation).



\subsection{Univariate Analysis}
We dig deeper into the outputs of our analyses to examine the impact of certain selected factors that were identified as important. The full set of results are presented in Appendix \ref{app:charts} and \ref{app:SIB}.

\medskip
\noindent \textbf{\em Impact of Pretraining Data Size (Train Token)} \quad Figure~\ref{fig:perf-TT} shows that for text classification using SIB-200, mBERT and XLM-R clearly obtain marked improvements as the language-specific pretraining data (train tokens) available to the models increases. To a lesser extent, this observation is also noticed for GPT-3. For text generation using mBBC, weaker relationship between performance and train tokens is observed for mBERT and GPT-3, while XLM-R fails to show any clear patterns.

\medskip
\noindent \textbf{\em Impact of General Resource Availability (Res Level)} \quad Figure~\ref{fig:perf-res} illustrates the performance of   \texttt{mBERT}, \texttt{XLM-R}, and \texttt{GPT-3}, across varying resource levels. For the text classification task on SIB-200, mBERT and XLM-R models perform similarly, while considerably outperforming GPT-3. However, in terms of trends related to resource levels, the results reinforce the significance of resource levels, with the lower resource levels (0, 1, and 2) showing weaker performance than relatively higher resource levels (3, 4, and 5), consistent across all three models. For the text generation task on mBBC, as expected it is \texttt{GPT-3}, the autoregressive model, that  performs much better than the MLM models \texttt{mBERT} and \texttt{XLM-R} models. However, for this task, the results are not as clearly distinct. While the highest resource level (5) continues to show a slight advantage over all the other levels, the gap is noticeably smaller. In other words, except for resource 5 level languages, increased resources do not necessarily guarantee improved performance. The results of languages in level 2 are often lower than those of 0 or 1, implying that the influence of resource availability on model performance is less pronounced in text generation task on mBBC.

\medskip
\noindent \textbf{\em Impact of Language Family} \quad Figures~\ref{figure:SIB-lang-res} and ~\ref{figure:lang+res} present the results of language family-based analysis on text classification and text generation tasks, respectively. In both the cases, we notice that  generally higher resource levels afford higher performance across all language families.  However, there is a considerable difference in the performance between the same resource levels but different language families, e.g., level 5 of Afro-Asiatic as compared to level 5 of Sino-Tibetan (Figure~\ref{figure:SIB-lang-res}) or level 3 of Austronesian as compared to level 3 of Dravidian or Indo-European (Figure~\ref{figure:lang+res}). While some of these differences may be in part due to the different number of languages present in each group, the results of this fine-grained analysis suggest that resource levels alone may not be sufficient indicator of performance. Moreover, the results of such a fine-grained analysis show no single language family as the most dominant feature.

These findings demonstrate the complex relationship between language families, resource availability, and model performance. While resource availability is important, other factors also influence performance within specific language families. 

\medskip
\noindent \textbf{\em Impact of Script Type} \quad Next, we analyze the impact of  script types on multilingual language model performance (Figure \ref{figure:script+res} in Appendix). One notable observation is that the  \texttt{GPT-3} model reveals a consistent superiority of the Latin script over other scripts in the text generation task. 

\section{Discussion}

Our study evaluates the performance of multilingual language models \texttt{mBERT}, \texttt{XLM-R}, and \texttt{GPT-3}. Some key observations can be summarized as follows:

\begin{itemize}[leftmargin=*]
\item Resource availability strongly correlates with model performance in text classification tasks but less so in text generation tasks. Instead, text generation on mBBC was influenced by factors such as language family and script type. 

\item The relationship between resource availability, language families, and model performance remains complex. While some language families exhibited consistent patterns across models, others showed varying results. 
Moreover, among the three models studied, there were notable differences, potentially due to their different training corpora. 

\item The impact of script type on model performance varied among the evaluated models. While \texttt{mBERT} and \texttt{XLM-R} showed no clear patterns between script types, the \texttt{GPT-3} model consistently performed better with the Latin script for text generation task. 


\end{itemize}

\section{Conclusion}

Our extensive evaluation of multilingual language models across two tasks consisting of 203 and 43 diverse languages, respectively, highlighted several interesting results. While certain models and tasks were impacted by resource availability (model-specific or general), language family and script types were found to be important factors for other models when used in another task. 
We plan to extend our research to incorporate newer large language models as well as explore the impact of additional factors, such as language-specific morphological features or syntactic structures, on model performance. 


\section*{Limitations}
Our study has several limitations that warrant acknowledgement. Firstly, the evaluation relied on two datasets, which may not fully encompass the diversity of languages and language usages. To obtain a more comprehensive understanding of multilingual language model performance, future work should incorporate additional datasets from diverse domains and genres.



Another limitation is the absence of fine-grained language identification and preprocessing steps in our data collection process when creating mBBC dataset. While this enabled direct retrieval of articles from specific news sources in each language, it may have introduced noise and inconsistencies into the dataset. Future research should consider integrating robust language identification and preprocessing techniques to enhance the quality and consistency of the dataset.


\section*{Ethics Statement}

The experimental setup and code implementation ensured adherence to ethical guidelines, data usage agreements, and compliance with the terms of service of the respective language models and data sources. The research team also recognized the importance of inclusivity and fairness by considering a diverse set of languages and language families in the evaluation, thereby avoiding biases and promoting balanced representation.


\bibliography{anthology,custom}

\appendix
\newpage
\section*{Appendix}

\section{\texttt{mBBC} overview} \label{app:mBBC}

\begin{figure*}[ht]
\centering
\includegraphics[width=0.7\textwidth]{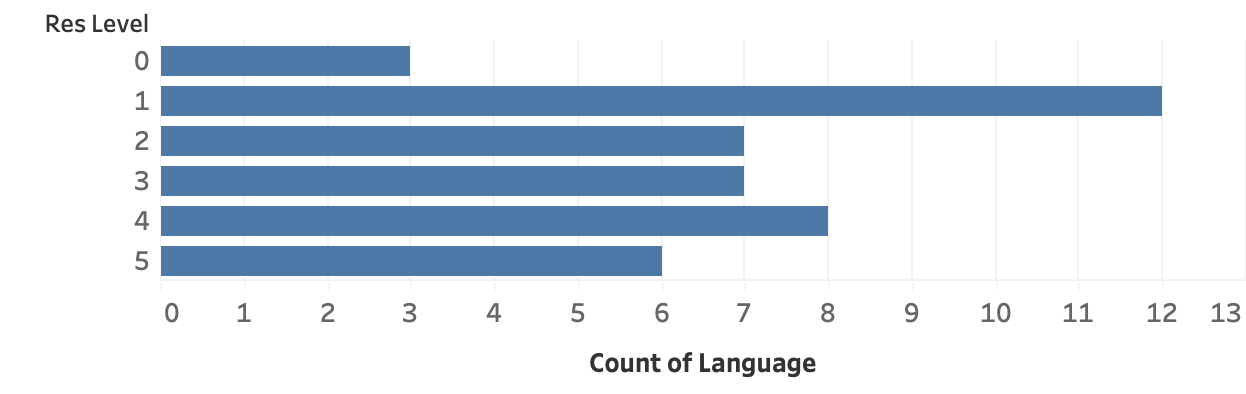}
\caption{Count of languages for each resources in \texttt{mBBC}.}
\label{figure:df-resources}
\end{figure*}

\begin{figure*}[ht]
\centering
\includegraphics[width=0.7\textwidth]{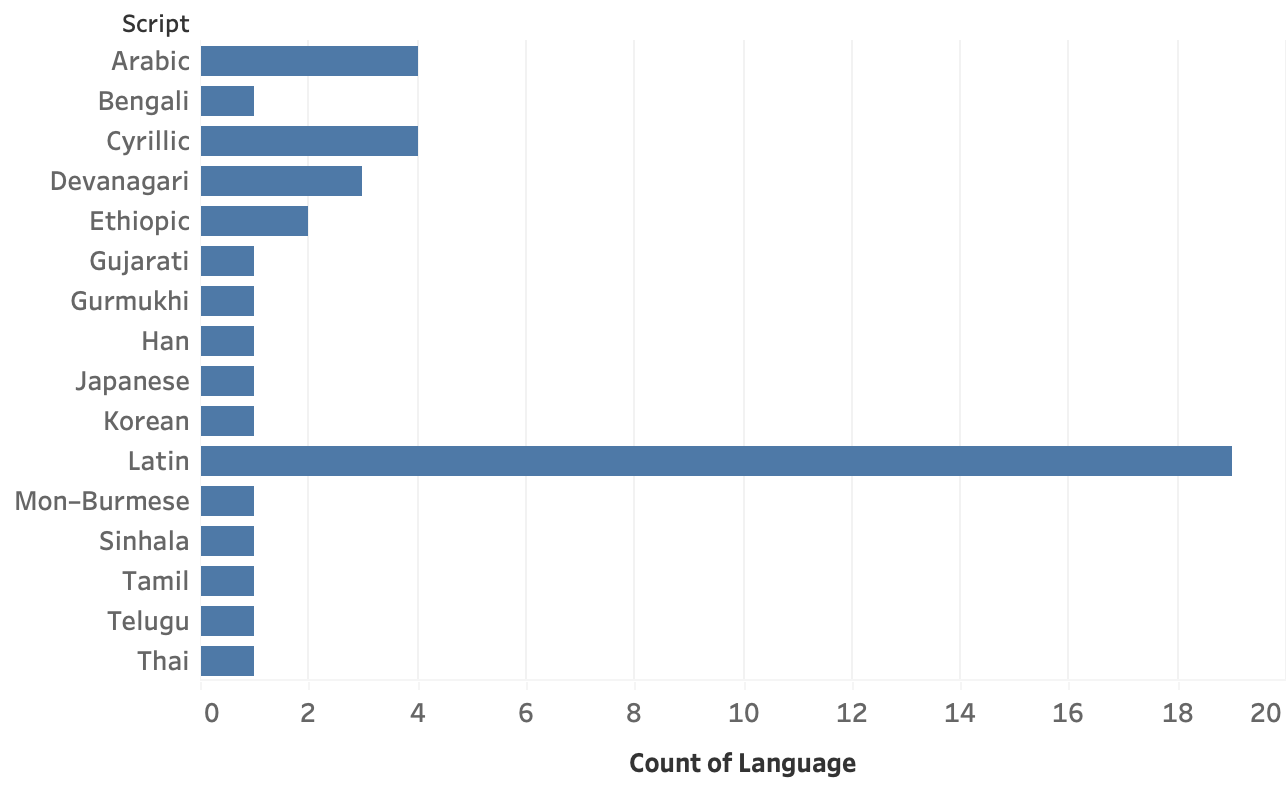}
\caption{Count of languages for each script in \texttt{mBBC}.}
\label{figure:ds-script}
\end{figure*}

\begin{figure*}[ht]
\centering
\includegraphics[width=0.7\textwidth]{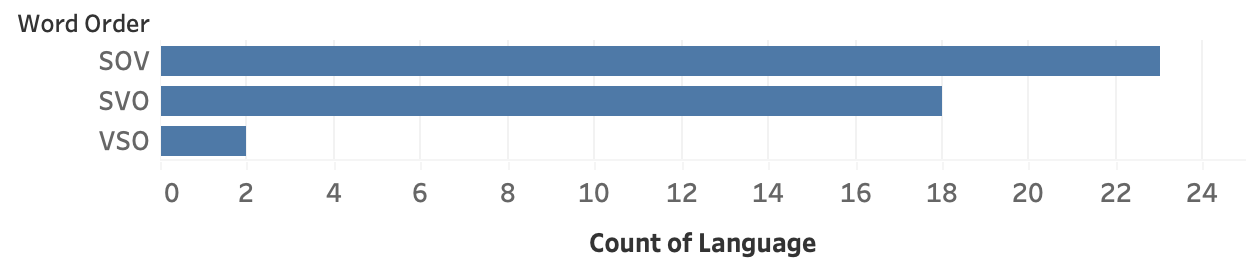}
\caption{Count of language for word order in \texttt{mBBC}.}
\label{figure:ds-wordorder}
\end{figure*}

Figures~\ref{figure:df-resources}, ~\ref{figure:ds-script}, and ~\ref{figure:ds-wordorder} present the distribution of languages in the \texttt{mBBC} dataset according to resource levels, language family, script, and word order, respectively.

\section{Experimental Procedure and Implementation} \label{app:procedure}
To evaluate the models, we followed a consistent experimental procedure across all languages:

\begin{enumerate}[leftmargin=*]
\item Select a language from the dataset and load the corresponding language model (e.g., bert-base-multilingual-cased for supported languages).
\item Randomly sample 2000 instances from the dataset for the chosen language.
\item For each instance, provide the model with an input sequence of 30 tokens and prompt it to predict the next token.
\item Rank the predicted tokens based on their probability scores.
\item Check if the ground truth token appears in the top 5 predicted tokens.
\item Repeat steps 1-5 for each language and model combination.
\end{enumerate}

\begin{table*}[t!]
\centering
\small
\begin{tabular}{l|l|l|c|c|c|c|c}
\toprule
\textbf{Language}                 & \textbf{Lang Family}    & \textbf{Script}        & \textbf{Word Order} & \textbf{\texttt{XLM-R} TT} & \textbf{\texttt{mBert} TT} & \textbf{\texttt{GPT-3} TT} & \textbf{Res Level} \\
\toprule
Afaan Oromoo             & Afro-Asiatic   & Latin         & SOV        & 8                  & 0                  & 0                 & 1         \\ \hline
Amharic                  & Afro-Asiatic   & Ethiopic      & SOV        & 68                 & 0                  & 0                 & 2         \\ \hline
French                   & Indo-European  & Latin         & SVO        & 9780               & 823                & 3553.1            & 5         \\ \hline
Hausa                    & Afro-Asiatic   & Latin         & SVO        & 56                 & 0                  & 0                 & 2         \\ \hline
Igbo                     & Niger-Congo    & Latin         & SVO        & 0                  & 0                  & 0                 & 1         \\ \hline
Kirundi (Rundi)          & Niger-Congo    & Latin         & SVO        & 0                  & 0                  & 0                 & 1         \\ \hline
Nigerian Pidgin          & Creole         & Latin         & SVO        & 0                  & 0                  & 0                 & 0         \\ \hline
Somali                   & Afro-Asiatic   & Latin         & SOV        & 62                 & 0                  & 0                 & 1         \\ \hline
Swahili                  & Niger-Congo    & Latin         & SVO        & 275                & 6                  & 0.6               & 2         \\ \hline
Tigrinya                 & Afro-Asiatic   & Ethiopic      & SOV        & 0                  & 0                  & 0                 & 2         \\ \hline
Yoruba                   & Niger-Congo    & Latin         & SVO        & 0                  & 1                  & 0                 & 2         \\ \hline
Kyrgyz                   & Turkic         & Cyrillic      & SOV        & 94                 & 11                 & 0.1               & 1         \\ \hline
Uzbek                    & Turkic         & Cyrillic      & SOV        & 91                 & 12                 & 1.5               & 3         \\ \hline
Burmese                  & Sino-Tibetan   & Burmese & SOV        & 71                 & 12                 & 1.9               & 1         \\ \hline
Chinese                  & Sino-Tibetan   & Han           & SVO        & 435                & 1551               & 193.5             & 5         \\ \hline
Indonesian               & Austronesian   & Latin         & SVO        & 22704              & 96                 & 116.9             & 3         \\ \hline
Japanese                 & Japonic        & Japanese      & SOV        & 530                & 713                & 217               & 5         \\ \hline
Korean                   & Koreanic       & Korean        & SOV        & 5644               & 81                 & 33.1              & 4         \\ \hline
Thai                     & Kra-Dai        & Thai          & SVO        & 1834               & 44                 & 26.8              & 3         \\ \hline
Vietnamese               & Austro-Asiatic & Latin         & SVO        & 24757              & 180                & 83.1              & 4         \\ \hline
Bengali                  & Indo-European  & Bengali       & SOV        & 525                & 42                 & 3                 & 3         \\ \hline
Gujarati                 & Indo-European  & Gujarati      & SOV        & 140                & 8                  & 0.5               & 1         \\ \hline
Hindi                    & Indo-European  & Devanagari    & SOV        & 1715               & 44                 & 9.4               & 4         \\ \hline
Marathi                  & Indo-European  & Devanagari    & SOV        & 175                & 11                 & 3.7               & 2         \\ \hline
Nepali                   & Indo-European  & Devanagari    & SOV        & 237                & 7                  & 1.1               & 1         \\ \hline
Pashto                   & Indo-European  & Arabic        & SOV        & 96                 & 5                  & 0                 & 1         \\ \hline
Punjabi                  & Indo-European  & Gurmukhi      & SOV        & 68                 & 12                 & 0.7               & 2         \\ \hline
Sinhala                  & Indo-European  & Sinhala       & SOV        & 243                & 0                  & 0.7               & 0         \\ \hline
Tamil                    & Dravidian      & Tamil         & SOV        & 595                & 42                 & 5.2               & 3         \\ \hline
Telugu                   & Dravidian      & Telugu        & SOV        & 249                & 41                 & 1.6               & 1         \\ \hline
Urdu                     & Indo-European  & Arabic        & SOV        & 730                & 22                 & 0.7               & 3         \\ \hline
Azerbaijani              & Turkic         & Latin         & SOV        & 783                & 36                 & 2.5               & 0         \\ \hline
English                  & Indo-European  & Latin         & SVO        & 55608              & 2623               & 181014.7          & 5         \\ \hline
Gaelic & Indo-European  & Latin         & VSO        & 21                 & 0                  & 0.8               & 1         \\ \hline
Russian                  & Indo-European  & Cyrillic      & SVO        & 23408              & 575                & 368.2             & 4         \\ \hline
Serbian                  & Indo-European  & Latin         & SVO        & 843                & 100                & 52.9              & 4         \\ \hline
Turkish                  & Turkic         & Latin         & SOV        & 2736               & 75                 & 116.1             & 4         \\ \hline
Ukrainian                & Indo-European  & Cyrillic      & SVO        & 6.5                & 263                & 14.9              & 3         \\ \hline
Welsh                    & Indo-European  & Latin         & VSO        & 141                & 16                 & 3.5               & 1         \\ \hline
Portuguese               & Indo-European  & Latin         & SVO        & 8405               & 312                & 1025.4            & 4         \\ \hline
Spanish                  & Indo-European  & Latin         & SVO        & 9374               & 689                & 1510.1            & 5         \\ \hline
Arabic                   & Afro-Asiatic   & Arabic        & SVO        & 2869               & 169                & 60.8              & 5         \\ \hline
Persian                  & Indo-European  & Arabic        & SOV        & 13259              & 106                & 16.7              & 4         \\
\bottomrule
\end{tabular}
\caption{\label{citation-guide}Linguistic Diversity in \texttt{mBBC} Dataset. 
The "\texttt{XLM-R} TT," "\texttt{mBERT} TT," and "\texttt{GPT-3} TT" columns represent the respective number of million tokens in the training dataset for each model. In the "Res Level" column, resource levels are indicated on a scale from 0 to 5, where 0 denotes an extremely low-resource setting, and 5 signifies a high-resource environment.
}
\label{table:dsview}
\vspace{-0.5cm}
\end{table*}

\begin{figure*}[ht]
\centering
\includegraphics[width=1\textwidth]{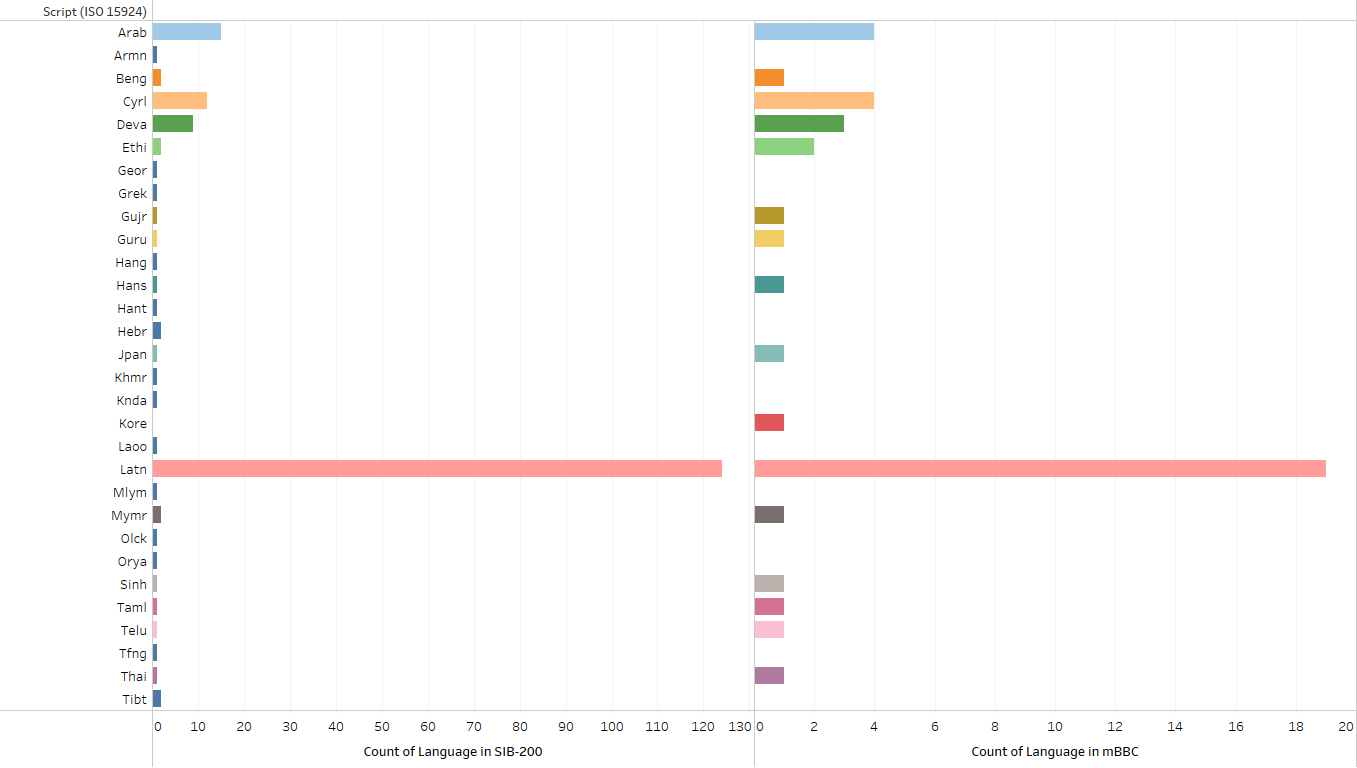}
\caption{Distribution of script in SIB-200 and \texttt{mBBC} datasets.}
\label{figure:script}
\end{figure*}

For our experiments, we leveraged the HuggingFace transformers library, a popular and flexible NLP library, to evaluate \texttt{mBERT} and \texttt{XLM-R}. This library offered a convenient and efficient framework for conducting experiments with different language models. To execute the experiments, we utilized Google Colab with a T4 GPU accelerator, enabling us to efficiently process a large number of samples across multiple languages and reduce overall processing time. For the \texttt{GPT-3} model, we employed the OpenAI API to implement and assess its performance in our tests.

\section{Additional Results for \texttt{mBBC} task} \label{app:charts}

\begin{figure*}[ht]
\centering
\includegraphics[width=1\textwidth]{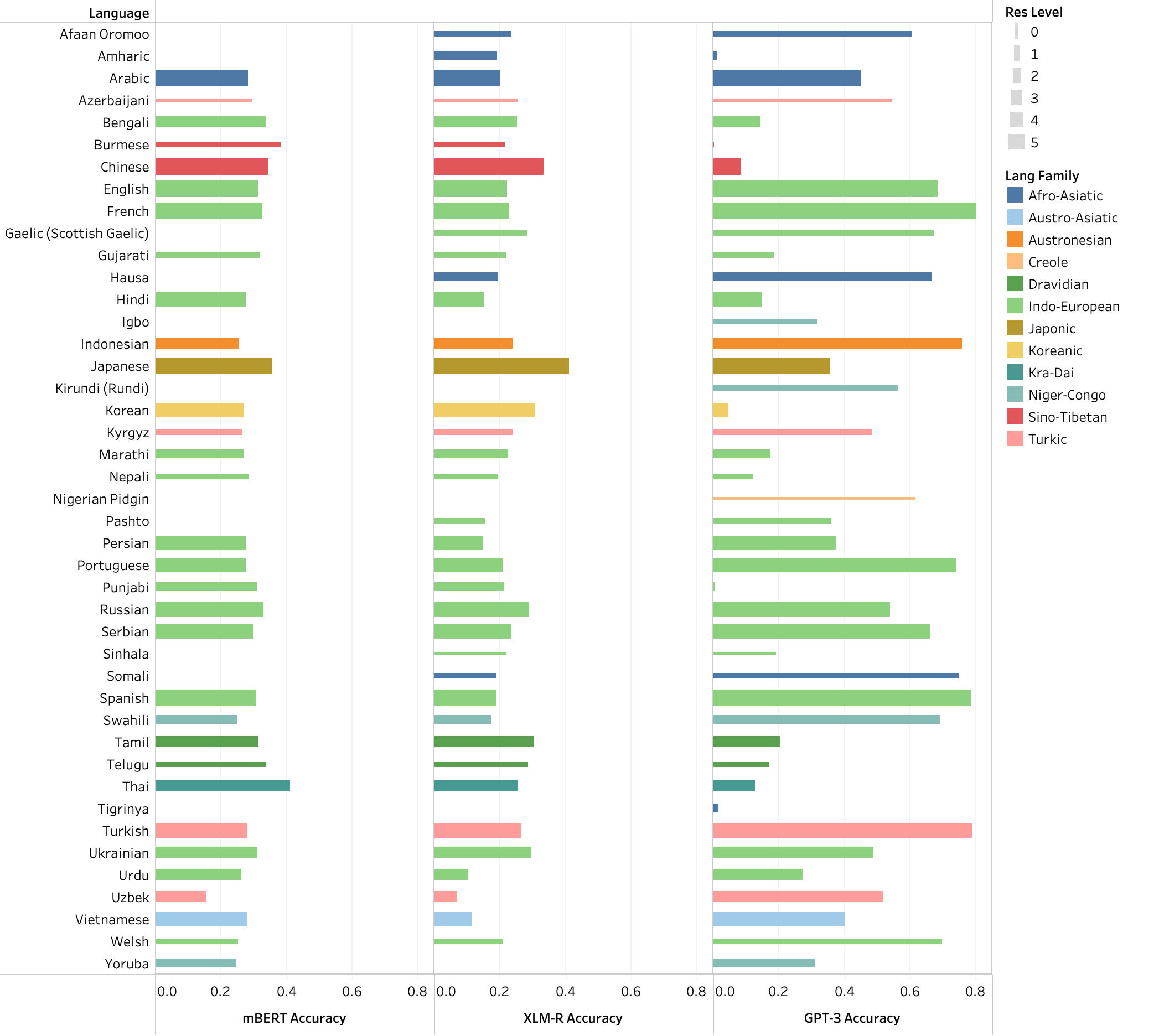}
\caption{Accuracy of \texttt{GPT-3}, \texttt{XLM-R}, and \texttt{mBERT} on \texttt{mBBC} for each language, with language family indicated by color and resources indicated by size}
\label{figure:lang-model}
\end{figure*}

The performance of each model in different languages, along with their respective language families and resource sizes, is depicted in Figure \ref{figure:lang-model}. This comprehensive visualization provides a clear overview of how each model performs across various languages and highlights the relationship between language characteristics and model performance.

\begin{figure*}[ht]
\centering
\includegraphics[width=1\textwidth]{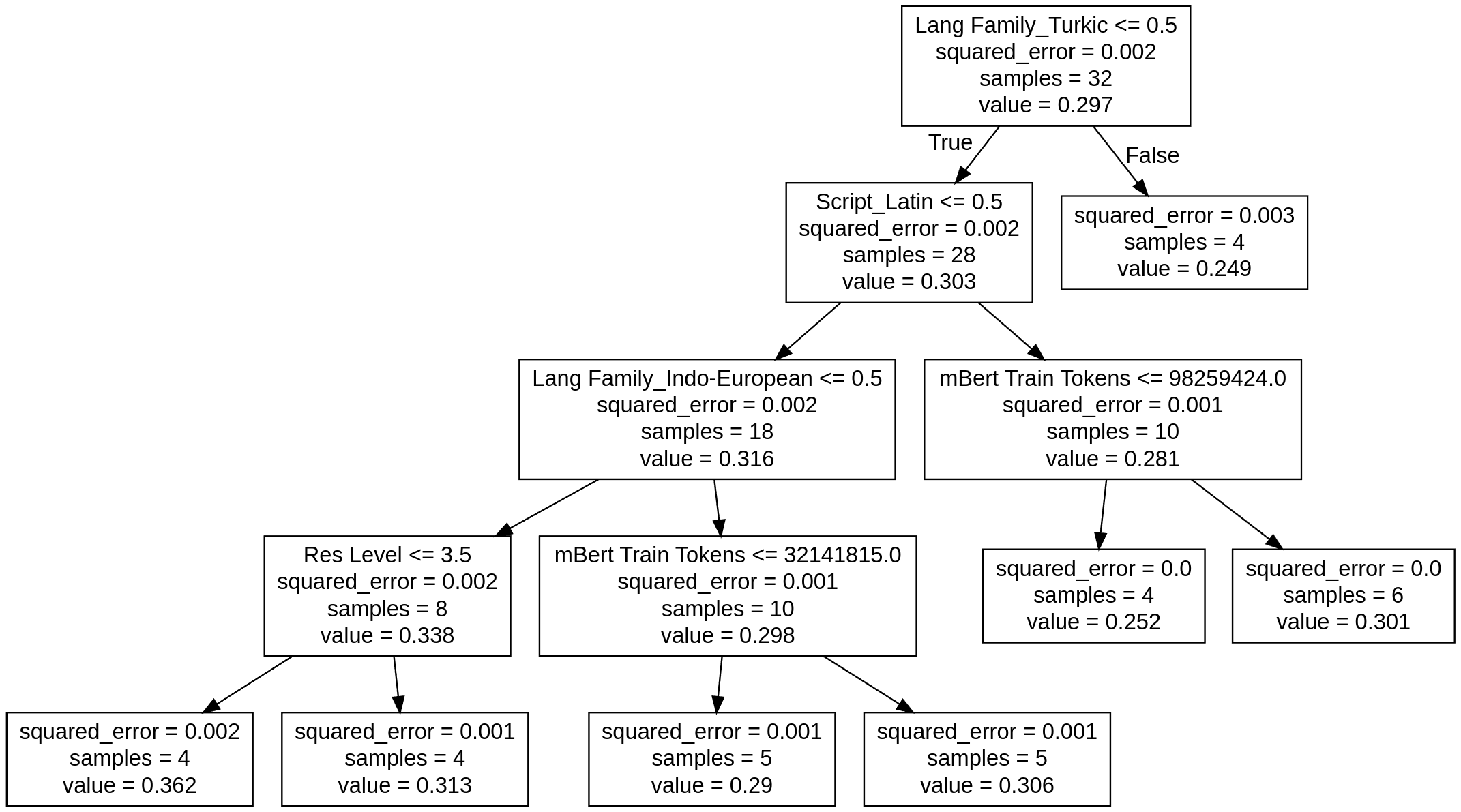}
\caption{Decision tree visualization for \texttt{mBERT} model on \texttt{mBBC} dataset.}
\label{figure:tree-mbert}
\end{figure*}

\begin{figure*}[ht]
\centering
\includegraphics[width=0.9\textwidth]{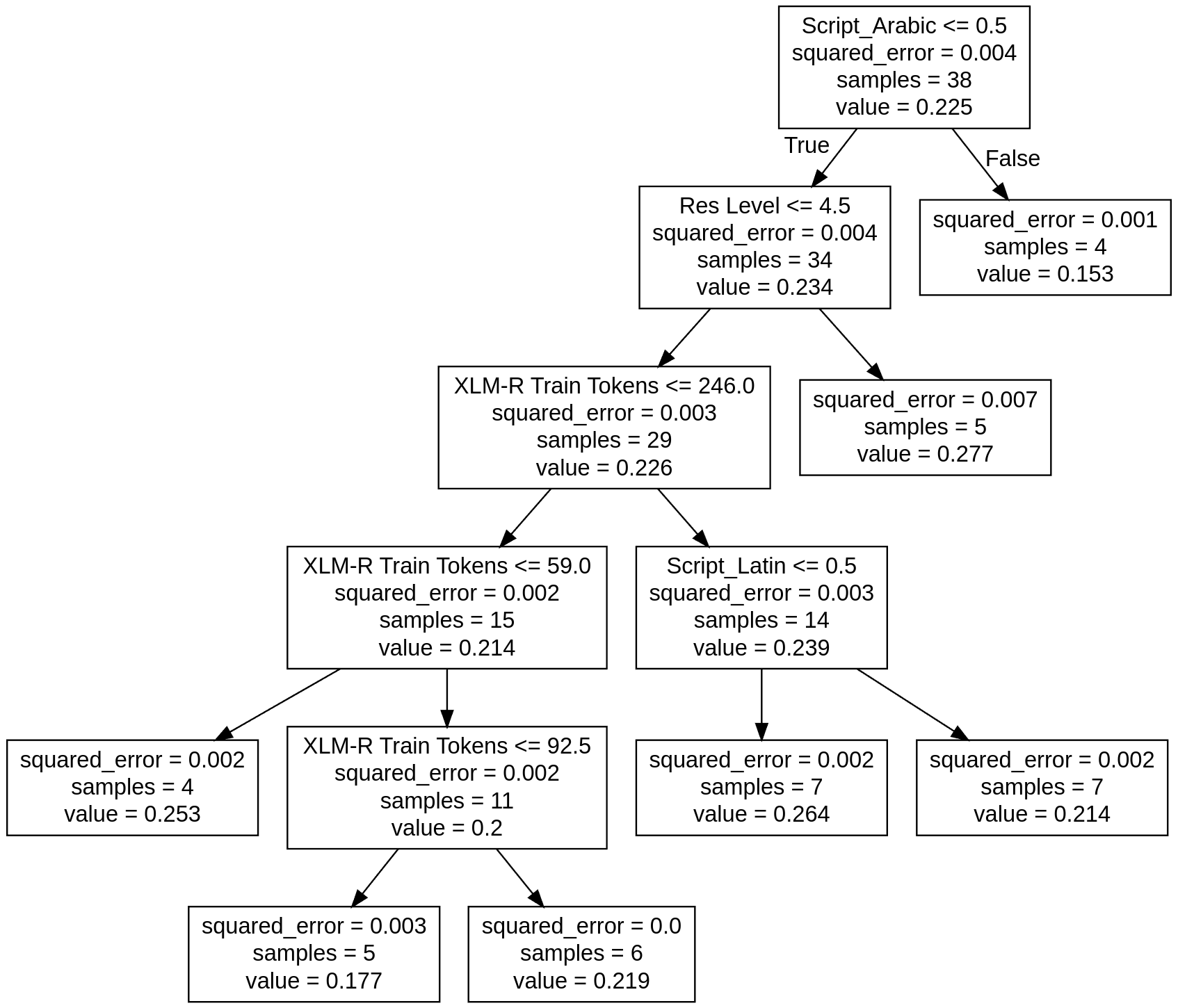}
\caption{Decision tree visualization for \texttt{XLM-R} model on \texttt{mBBC} dataset.}
\label{figure:tree-xlmr}
\end{figure*}

To gain deeper insights into the decision-making process of each model, Figures \ref{figure:tree-mbert} and \ref{figure:tree-xlmr} present the complete decision trees for \texttt{mBERT} and \texttt{XLM-R}, respectively. These decision trees provide a detailed representation of the factors influencing the models' predictions, offering a comprehensive view of the underlying mechanisms employed by each model.

\begin{figure*}[ht]
\centering
\includegraphics[width=0.75\textwidth]{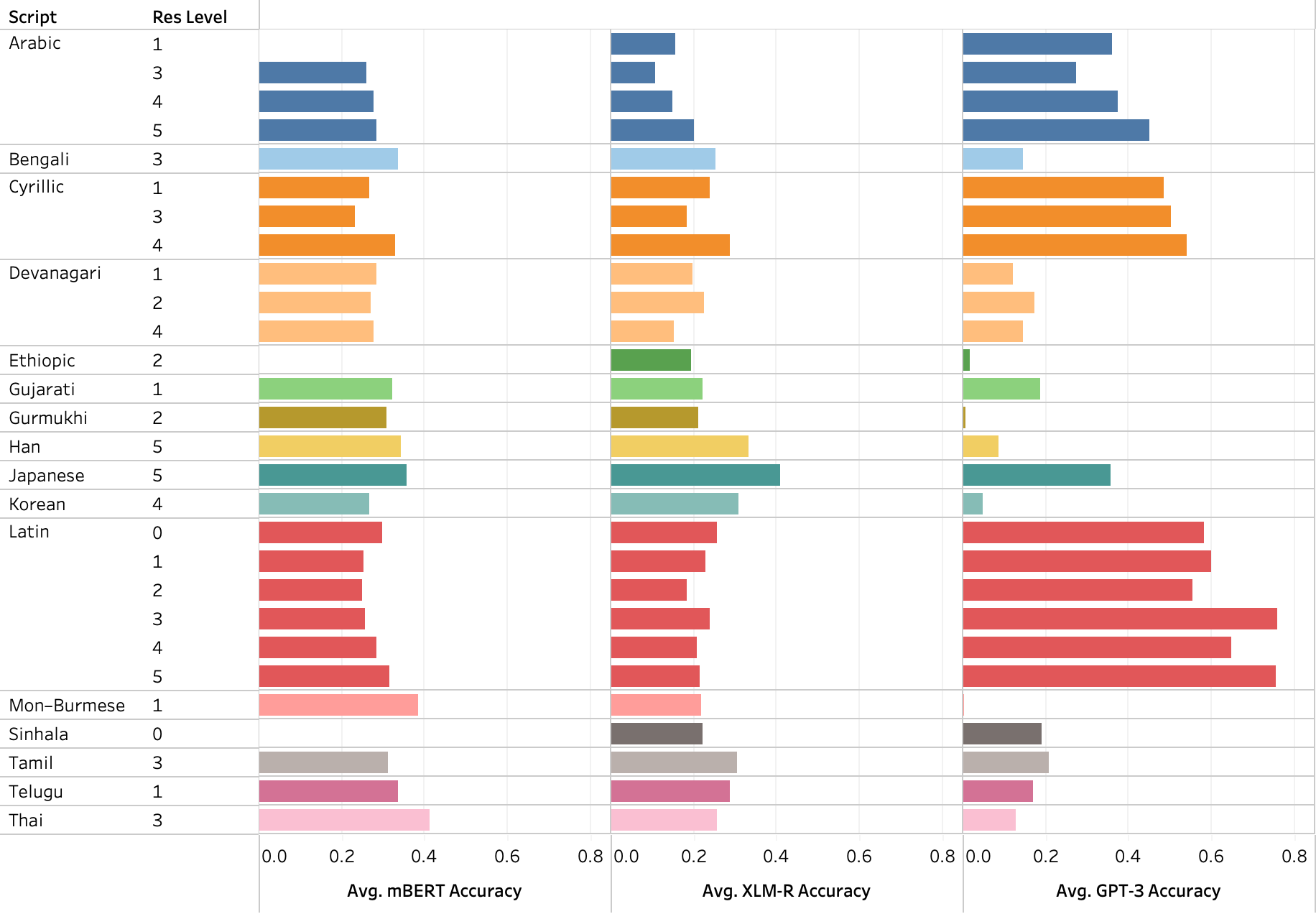}
\caption{Average accuracy of \texttt{GPT-3}, \texttt{XLM-R}, and \texttt{mBERT} for each script on \texttt{mBBC}}
\label{figure:script+res}
\end{figure*}

\FloatBarrier
\section{SIB-200 Downstream Task Results} \label{app:SIB}
\begin{figure*}[ht]
\centering
\includegraphics[width=1\textwidth]{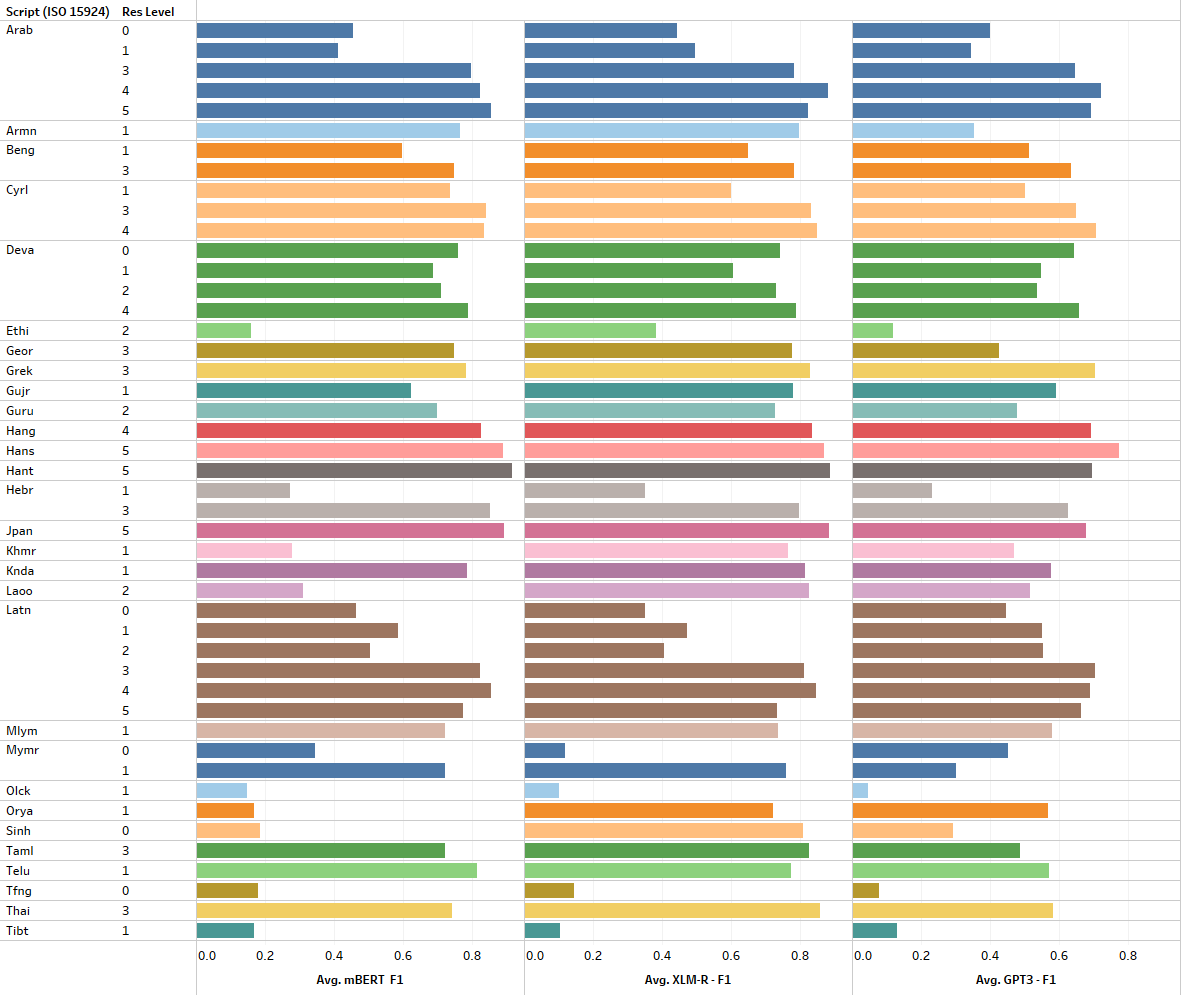}
\caption{Average accuracy of GPT-3, XLM-R, and mBERT for each script in SIB-200 task.}
\label{figure:SIB-scri-res}
\end{figure*}

In this appendix, we provide detailed results from the downstream task evaluation on the SIB-200 dataset. The Figure \ref{figure:SIB-lang-res} and \ref{figure:SIB-scri-res} present the average F1 scores, of \texttt{mBERT}, \texttt{XLM-R}, and \texttt{GPT-3} on the text classification task in each of the different language models and scripts covered by the SIB-200 dataset. These results offer a comprehensive overview of how each model performed on the specific classification task, highlighting variations in performance across different languages. Additionally, we include decision tree analyses between language features, such as language family, script type, and resource availability, and the F1 scores achieved by each model. These trees provide insights into the factors influencing model performance in the context of the SIB-200 downstream task and Mann-Whitney U test proved the effect of selected features on F1 score. The detailed results and analyses presented in this appendix contribute to a thorough understanding of the language models' capabilities in addressing diverse linguistic challenges in practical applications.

\begin{figure*}[ht]
\centering
\includegraphics[width=1\textwidth]{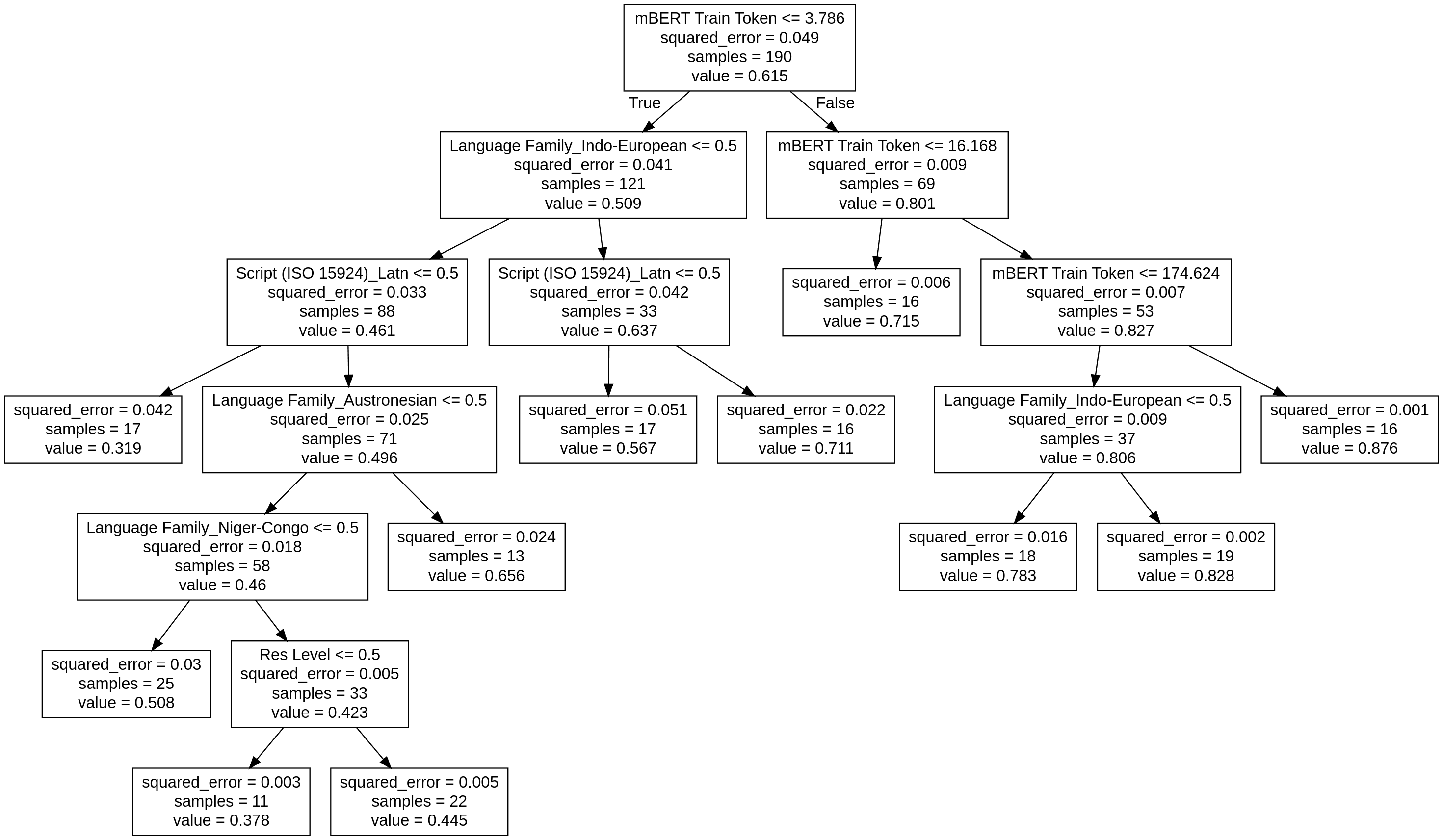}
\caption{Decision tree visualization for \texttt{mBERT} model on SIB-200 dataset}
\label{figure:tree-SIB-mBERT}
\end{figure*}

\begin{figure*}[ht]
\centering
\includegraphics[width=1\textwidth]{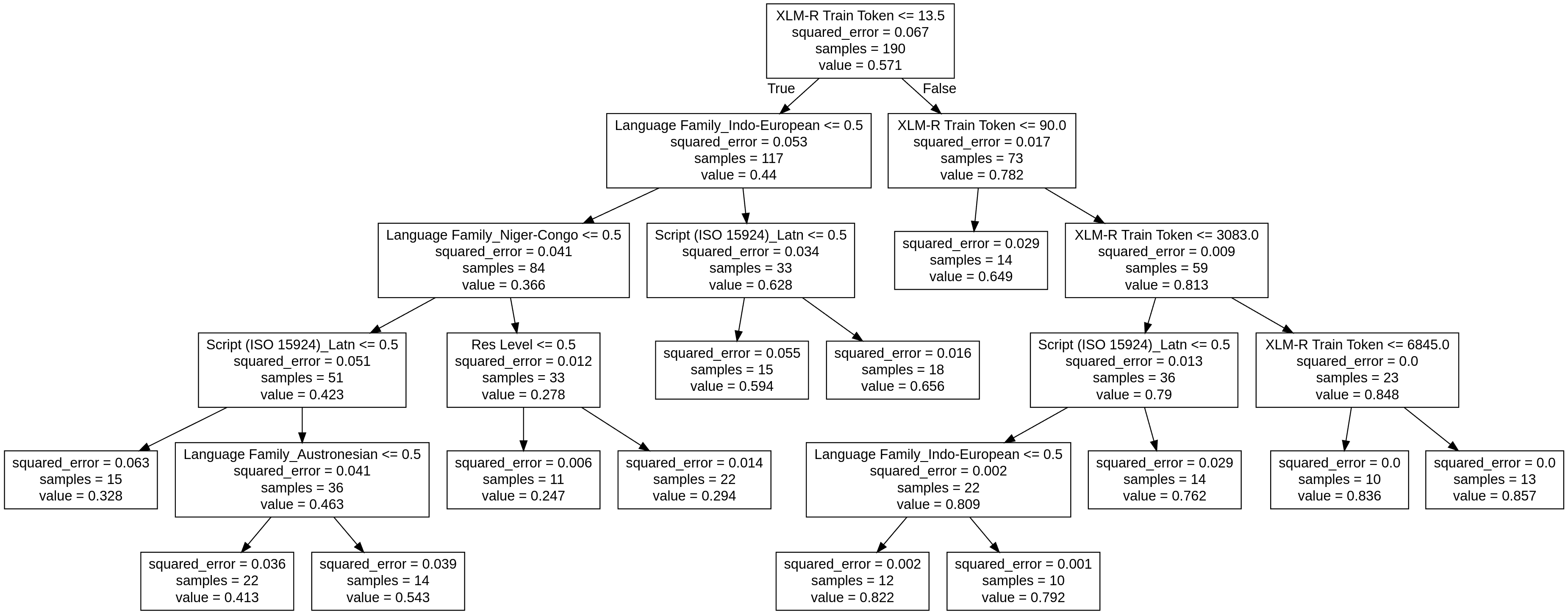}
\caption{Decision tree visualization for \texttt{XLM-R} model on SIB-200 dataset}
\label{figure:tree-SIB-XLMR}
\end{figure*}

\end{document}